\documentclass[10pt,letterpaper]{article}

\usepackage{cogsci}
\usepackage{pslatex}
\usepackage{apacite}
\usepackage{float} % Roger Levy added this and changed figure/table
\title{Beyond Transformers for Function Learning}
 
\author{{\large \bf Simon Segert (ssegert@princeton.edu)} \\
  Princeton Neuroscience Institute, Washington Road \\
  Princeton, NJ 08544 USA
  \AND {\large \bf Jonathan Cohen (jdc@princeton.edu)} \\
  Princeton Neuroscience Institute, Washington Road \\
  Princeton, NJ 08544 USA}

\begin{document}

\maketitle

\begin{abstract}
The ability to learn and predict simple functions is a key aspect of human intelligence. Recent works have started to explore this ability using transformer architectures, however it remains unclear whether this is sufficient to recapitulate the extrapolation abilities of people in this domain. Here, we propose to address this gap by augmenting the transformer architecture with two simple inductive learning biases, that are directly adapted from recent models of abstract reasoning in cognitive science. The results we report demonstrate that these biases are helpful in the context of large neural network models, as well as shed light on the types of inductive learning biases that may contribute to human abilities in extrapolation.

\end{abstract}

\section{Introduction}
\label{introduction}
People often reliably identify patterns or rules in small amounts of data, and extrapolate them to domains out of the range of the training data. Despite its seeming simplicity, this process has proved deceptively hard for most neural networks, which often require orders of magnitude more training data.  The extrapolation of scalar functions has also been used as an informative test case for abstract reasoning for both humans \cite{delosh} and neural networks \cite{tmlr} alike.  

Recently, the transformer architecture has shown promising abilities to interpolate simple scalar functions \cite{Oswald,garg}. However, it remains unclear whether such models can match human abilities on the more challenging and informative task of \textit{extrapolating} simple, scalar functions. Moreover, to the extent that they fall short in doing so, an important question is: what kinds of modifications or inductive biases would improve their performance? In particular, can ideas from cognitive science be scaled up to aid performance of models using this architecture?

In this work, we propose two such learning biases that are heavily adapted from previous cognitive models, and that we hypothesize would lead to such an improvement. We term them ``Relational Bottleneck" and ``Adaptive Attentional  Window" . The first has been used in models of abstract rule learning, as a way to enforce a separation between content and relations \cite{esbn,corelnet}. The second has been proposed in the context of a cognitive model of function learning \cite{maxent}, and may be thought of as the assumption that the value of a function at a given timepoint depends only the values at a small number of nearby timepoints, with the exact length scale needing to be learned. 
%(e.g., its fundamental domain (??RELEANT CITE?)

Our contribution is first to show how each of these biases can be implemented in the transformer architecture, and secondly to show that their presence improves the performance of this model on extrapolation of the sorts of scalar functions that are easy for people. 

Finally, as a collateral benefit, we show that the method we use to implement the relational bottleneck yields \textit{uncertainty estimates} without any additional computation, which standard function learning transformers (such as in \cite{garg}) do not provide.

The remainder of the paper is structured as follows: We first provide background and further explanation of the two learning biases.  We also describe how we implemented these in the context of a transformer architecture, and how the resulting architecture can also be used to naturally model uncertainty. 
We then empirically evaluate the resulting model and compare it with a standard transformer. Finally we conclude with an overview of related work and general discussion. 

\section{Background}

\subsection{Linear Autoregressive Function Learning-Motivating Adaptive Attention Window}
\label{window_motivation}
Function learning, that is, the study of peoples' abilities to interpolate and extrapolate scalar functions from a small number of observations, is a classic topic in cognitive psychology \cite{delosh,mcdaniel}. The most popular models of peoples' function learning abilities are based on Gaussian Processes \cite{lucas, schulz,wilson}. While this framework is able to explain many specific phenomena, it has the disadvantage of requiring pre-specification of a class of functional or distributional forms, and also makes some predictions that are qualitatively at odds with human behavior, especially in cases where there is not a simple parametric underlying pattern to the data \cite{maxent}. It has subsequently been proposed that these issues can be addressed using linear autoregressive models (ibid.) In this case, the value of the function at a location $i\in\mathbf{N}$ is assumed to be a linear combination of a small number of preceding values: $f(i)=\sum_{j=1}^L w_jf(i-j)$, with the parameters $w_j$ fit to a given function, and the ``window length" $L$ being a hyperparameter. A crucial feature of this model is the assumption of dependence of $f(i)$ on only a small number of nearby values of $f$. This property is psychologically appealing, as it naturally maps onto constraints on human memory or processing. It is also mathematically convenient, since it limits the total number of parameters $w_j$ that must be fit. Due to the success of this simple and general model at fitting human extrapolation performance, we hypothesize that an adaptation of the finite window length will be beneficial for transformers in function learning tasks as well. 

\subsection{Systematic Generalization with Neural Networks-Motivating Relational Bottleneck}
\label{simmat_motivation}
While the first bias was motivated by a cognitive model of function learning, use of a relational bottleneck was motivated by models of abstraction and relational reasoning \cite{esbn,corelnet}, and a correspondence between such tasks and function learning. To see this, we first consider that a simple function such as a line can be thought of as a kind of simple abstract \textit{rule} (e.g., increment a fixed amount for each segment) or, alternatively, a relation between each point and the next. The aforementioned works have considered how to design neural networks to systematically extract and apply abstract rules in visual reasoning settings, such as detecting whether two images are the same or different. These models have the ability to learn a rule on a given collection of inputs (e.g., stars) and then apply the same rule to a different set of unseen images (e.g., circles). At a high level, such models work by constructing a similarity matrix between encoded representations of objects, and forcing the remainder of processing to be done on this pattern of similarities. In this way, the part of the model responsible for learning and applying the rule do not (by design) have access to ``sensory" details of specific objects, and thus are forced to learn and use second-order relational patterns. We refer to this as the Relational Bottleneck assumption. Due to the success of these models, we hypothesize that the inductive bias of focusing on the pattern of relations between the values of a function, rather than the raw values themselves, will also aid transformers in function learning tasks.

To give a bit more detail on the models,
the ESBN model \cite{esbn} uses a mechanism of binding and similarity-based retrieval in an external memory, in order to force a reasoning module to operate only on patterns of similarities. The ESBN mechanism has also been adapted for another simple form of extrapolation, namely counting with integers \cite{dulberg}. This further suggests that it may be possible to also adapt it to the case of extrapolating smooth function as well (after all, an increasing linear function is very similar to a sequence of consecutive integers).

The CoRelNet model \cite{corelnet} can be regarded an abstraction of ESBN model, that implements the bottleneck in a simpler and more explicit form. This model takes a sequence of objects as input, and runs each through a shared encoder module, to obtain representations $z_i\in\mathbf{R}^d$ for each object $i=1,\dots, n$. These are used to construct a representational similarity matrix $S_{ij}=<z_i,z_j>$. Finally a reasoning module (architecturally an MLP) is used to read out the answer from the similarity matrix.

\section{Transformers and Function Learning}
\label{transformer_background}
Finally, we give a brief overview of the Transformer architecture \cite{vaswani} as we employ it. As in previous work on modeling sequences, we will employ a decoder-only transformer architecture \cite{radford}. 
At the specification level, the transformer architecture $TF$ takes as input a list of vectors $\{v_i\}_{i=1}^N$ and returns as output a list consisting of a ``transformed" version of each input vector: $TF(\{v_i\}_{i=1}^N)=\{\tilde{v_i}\}_{i=1}^n$. Some recent works have used this architecture as a model for function learning \cite{Oswald,garg}. In these works, the input is a sequence of ordered-pairs $\{(x_i,y_i)\}_i$ of observations of the values of a function $f$, together with a query point of the form $(x_q,0)$, where $x_q$ is a point at which the value of the function is to be estimated. The set of such points are passed through a transformer, and the transformed vector corresponding to the query point is finally passed through a learned linear decoder in order to obtain an estimate for $f(x_q)$. 

Since the details of the architecture are described in many other papers (e.g. \cite{vaswani}) we will instead highlight  two main features that are particularly relevant to our purposes. The first is \textit{permutation equivariance}, meaning that if the input vectors are re-ordered according to some permutation, then the output vectors will also be reordered in the same way. Symbolically, $TF(\{v_{\sigma(i)}\}_{i=1}^N)=\{\tilde{v}_{\sigma(i)}\}_{i=1}^n$, where $\sigma$ is any permutation. This property means that the input to the transformer can be regarded as an \textit{unordered set}, rather than as an ordered list. This is important because it means that the input does not have to be a one-dimensional sequence, which property we will exploit subsequently. The second property is the presence of masking. When computing self-attention weights between elements in the input sequence, it is common to impose a value of 0 on the weight between certain pairs. This is often done, for example, when the inputs have a canonical temporal ordering, and one does not want to allow one input to attend to those that occur in the future \cite{radford}. This process can be regarded as a special case of multiplicative weighting of the attention weights, where each gating factor is either 0 or 1. We exploit this in the next section when discussing the Adaptive Window.

\section{Methods}
\label{methods}
As indicated above, we aim to incorporate two inductive biases from the cognitive science literature into a transformer, to determine whether these improve its ability to extrapolate: (1)Relational Bottleneck, and (2) Adaptive Attention Window. We begin by describing the form of the similarity metric we use for the relational bottleneck (tailored to the context of function learning), followed by our implementation of the two inductive biases in the context of the transformer architecture.

In what follows, we will represent a scalar function as a list of x-y observations: $\{(x_i,y_i)\}_{i=1}^N$ where $x_i,y_i\in\mathbf{R}$ Furthermore, we will assume for the sake of simplicity, and following \cite{maxent}, that the x points are equally spaced along the horizontal axis: $x_i=i, i=1,\dots, N$. 

\subsection{Form of Similarity Measure}
\label{simmat_form}
In the ESBN and CoRelNet, the form of the similarity is given by either a cosine similarity or dot product similarity between the vectors. However, our case differs in that we are dealing with scalar functions (that is, each value of the function lives in a one-dimensional vector space), and we want to consider the similarity between the values of the function at different time points. The cosine and dot product similarity metrics have undesirable properties for such one-dimensional spaces, which make them unsuitable for our application. More specifically, the cosine similarity is degenerate in this context, which follows directly from the definition ${\frac {\langle v,w\rangle} {\|v\|\|w\|}}$ that the cosine similarity of any two non-zero one-dimensional vectors will be equal to 1. The dot product similarity $\langle v,w\rangle$ is also not suitable for one-dimensional vectors, due to its dependence on the scale of the values. For example, the dot product similarity between two small numbers is smaller than between a large number and a much larger number, even if the two small numbers are extremely close to each other while the two large numbers are very far from each other.  For these reasons, we use negative Euclidean distance as our measure of similarity, so that the (dis)similarity between two observations is defined as 
$$S(x,y)=x-y$$
where values with absolute value near zero correspond to very similar observations, and those with large absolute values correspond to dissimilar ones.  We choose a signed rather than unsigned distance because this will make it easier to translate from a similarity value to a raw function value, as we discuss further in the next section. 

\subsection{Incorporating Similarity Matrix Bottleneck}
\label{simmat_implementation}

As discussed in the previous section, the critical inductive bias in the ESBN and CoRelNet architectures is the imposition of an architectural constraint to allow the "reasoner" in the model to see only similarity scores between pairs of inputs, rather than raw input values.  Accordingly, in the case of predicting a scalar function, we restrict the reasoning component of our network, rather than seeing the raw values $\{y_i\}_{i\leq N}$ of the function, to see only the similarity matrix $\{S(y_i,y_j)\}_{i\leq N,j\leq N}$, with $S$ as defined in the previous section. Having specified the desired input, we now consider the desired target. We argue that, given that we restrict the network to consider only relational information, it would defeat the purpose if the prediction target was the raw value $y_{N+1}$. What, then, should the network predict? We posit that the natural prediction target, given the desideratum of imposing a Relational Bottleneck, is for the network to predict the \textit{N+1 st row of the similarity matrix}. Fortunately, this can easily be accomplished using a standard transformer architecture. We exploit the fact mentioned in the transformer description above that the input to a transformer can be an arbitrary set. We simply construct a set which consists of all values of the similarity matrix, together with their positions within the matrix, as well as a set of query points that encode the locations at which we want the model to predict the similarity values. That is, given an input $\{(x_i,y_i)\}_{i=1}^N$, we construct the following set:
$$X=\{(x_i,x_j,S(y_i,y_j))\}_{i<j\leq N}\cup \{(i,N+1,0)\}_{i\leq N+1}$$

Due to (anti)symmetry of the similarity function, it is enough to consider only the lower triangular part of the matrix $i<j$. This also helps to alleviate memory demands of the implementation.  This set is first passed through a learned linear projection into $\mathbf{R}^{d_{model}}$, and then fed into a transformer. The outputs of the transformer corresponding to the elements $(i,N+1,0)$ are finally passed through a learned linear decoder in order to obtain predictions for the next row of the matrix. This model is trained by minimizing the mean-square-error of the output relative to the vector $\{Sim(y_i,y_{N+1})\}_{i=1,\dots, N+1}$.  We refer to this model as the \textit{Relational Transformer}.

\subsection{Incorporating Adaptive Attention Window}
\label{window_implementation}
Following the autoregressive linear function learning model from \cite{maxent}, as well as the general transformer architecture, we map the window length $L$ to the transformer implementations simply by masking out any attention weights between observations that are separated by a distance $>L$ on the x-axis. However, in order to facilitate gradient-based learning, we do not actually impose such a hard-cutoff, but rather parameterize a multiplicative mask on input-to-input attention weights, using a learnable monotonic function of the distance between the inputs, thus allowing the network to learn the most effective interaction length scale. We refer to this as a \textit{Learned Attention Window}. 

We first explain the implementation of the Learned Window for the case of a standard transformer model, such as in \cite{garg}, that takes as input $\{(x_i,y_i)\}_{i=1}^N\cup \{(x_{N+1},0)\}$ and is trained to predict $y_{N+1}$. In this case, we would impose a multiplicative gating on the corresponding attention self-attention weight between $(x_i,y_i)$ and $(x_j,y_j)$, by a factor of $F_{\theta}(|x_i-x_j|) \textbf{1}_{x_i>x_j}$, where $F$ is a positive decreasing function such that $F(0)=1$, with $\theta$ being learnable parameters. Here the indicator function $\textbf{1}_{x_i>x_j}$ enforces the standard causal constraint on mask values \cite{radford}. Note that a gating of only $\textbf{1}_{x_i>x_j}$ would correspond to the standard upper-triangular mask. We parameterize $F$  using a decreasing sigmoidal form, $F_{a,b}(x)={\frac {1-\sigma(x/b-a)}{1-\sigma(-a)}}$, $a,b>0$,$\sigma(x)={\frac 1 {1+e^{-x}}}$ although other choices are possible. 

The implementation of the Adaptive Window for the Relational Transformer is similar. In this case, the input to the model is instead a set of tuples of the form $(x_i,x_j,S(y_i,y_j))$. Given two such tuples $(x_i,x_j,S(y_i,y_j))$ and $(x_{i'},x_{j'},S(y_{i'},y_{j'}))$, we impose a multiplicative gating factor on the self-attention weight between them as follows:
$$F_{\theta}(|x_i-x_{i'}|)F_{\theta}(|x_j-x_{j'}|)\textbf{1}_{i>i'}\textbf{1}_{j>j'}$$.  In this way, the gating factor decays with both horizontal and vertical distance within the similarity matrix. Furthermore, the indicator functions enforce a constraint similar to the standard causality constraint in the one-dimensional case. In that case, each element in a sequence is only allowed to attend to elements to the left of itself, whereas in the relational case, each entry in the similarity matrix is only allowed to attend to elements to the left and above itself.

\subsection{Function Extrapolation with Relational Transformer}
\label{extrapolation_implementation}
Given a set of values $\{(x_i,y_i)\}_{i=1}^N$, the Relational Transformer model predicts the ``similarity profile" of $y_{N+1}$, that is to say, the vector containing the similarity of $y_{N+1}$ with all preceding $y_i$. However, at test time what we want is $y_{N+1}$ itself. In order to recover this value from the predicted similarity profile, we exploit the fact that the similarity function has a known and simple mathematical form, namely an arithmetic difference.

Let $\hat{z}_{N+1}$ denote the predicted similarity profile. By definition, the \textit{i}-th component $(\hat{z}_{N+1})_i$ is the model's prediction of $S(y_{N+1},y_i)$. 
That is,  $(\hat{z}_{N+1})_i$ is the model's estimate of $y_{N+1}-y_i$. Thus by simply adding $y_i$, we can convert this estimate of $y_{N+1}-y_i$ into an estimate of $y_{N+1}$ itself. Doing this for each $i\leq N$, we obtain an ensemble of estimates $(\hat{y}_{N+1})_i$ for $y_{N+1}$, defined by

$$(\hat{y}_{N+1})_i=(\hat{z}_{N+1})_i+y_i$$

Note that we have used the \textit{invertibility} of the similarity function in a key way (more precisely, of the function $S(x,\cdot)$ for any $x$). This is why we used signed distance rather than euclidean distance when defining the similarity function. We return to this point in the discussion section.

Thus the output of the Relational Transformer model can be modified to yield an ensemble of estimates for $\hat{y}_{N+1}$. In turn, given an ensemble, we can naturally give both a point estimate and an uncertainty estimate for $y_{N+1}$. We define the point estimate as 
$$\hat{y}_{N+1}=Median_{i\leq N} (\hat{y}_{N+1})_i$$. 

The uncertainty estimates are treated similarly, with the uncertainty in the estimate being defined as the sample standard deviation of the ensemble $(\hat{y}_{N+1})_i, i\leq N$. 

\subsection{Comparison model}
\label{comparison}
Since our objective is to understand what inductive biases are useful to transformers, we will consider as a control a model following \cite{garg}. In this model, we first construct a set $\{(x_i,y_i)\}_{i=1}^N\cup \{(x_{N+1},0)\}$, and then pass each vector in the set through a shared learned linear embedding to $\mathbf{R}^2\to \mathbf{R}^{d_{model}}$. The resulting set of vectors is then fed through a transformer, and the output vector corresponding to the token $(x_{N+1},0)$ is passed through a learned linear decoder $\mathbf{R}^{d_{model}}\to\mathbf{R}^1$ to obtain the model's prediction of $y_{N+1}$. This model is trained using mean-square-error of the prediction relative to the true value $y_{N+1}$. We will sometimes refer to this as the 1d Transformer model, to distinguish it from the Relational Transformer model above.

\section{Experiment details}
\label{experiment_details}
Models were trained on a next-timestep prediction objective using squared error loss as described in the previous section. 

In our experiments we consider functions of length $N=20$, which is similar to values used in similar psychological experiments, e.g. \cite{ciccione}. The training data for each model consisted of a combination of lines, sinusoids, and Radial Basis Function (RBF) curves, following previous works such as \cite{schulz,tmlr}. 

The lines were sampled with slopes in $[-.1,.1]$. The sinusoids were sampled with periods in $[5,12]$, amplitudes in $[.8,1.2]$ and phases in $[0,2\pi]$. The RBF curves were sampled from a Gaussian distribution with mean $0$ and covariance $C_{ij}=e^{-.5*(i-j)^2/\sigma^2}$, where  $\sigma=3$. When sampling training curves, each of the above three classes was sampled with probability $1/3$, and then the parameters within each class were sampled as described above in order to generate the actual curve. Additionally, all curves had random uniform noise added to the y values with mean 0 and $\sigma=.1$.

Both the Relational Transformer and one-dimensional transformer models have an embedding dimension of $d_{model}=256$, with 8 attention heads and 12 layers. We do not use dropout. 

All models were trained on a total of 320000 curves using a batch size of 32. We used the Adam optimizer with default parameters and learning rate of $10^{-4}$.  Each model was trained 3 times from different random initializations. All simulations were done using PyTorch.

\section{Results}
\label{results}
In table \ref{tab:mse_results_table}, we consider extrapolation results on the three classes of curves. We sampled a total of 2500 new curves evenly split among the three classes linear,sine, and rbf. We used each model to extrapolate the function to the points $x_{N+1},\dots, x_{N+10}$ in an autoregressive fashion, similar to \cite{radford}. That is, after we have obtained the model's prediction $\hat{y}_{N+1}$ for the value at $x_{N+1}$, we construct a new input set consisting of the original observations $\{(x_i,y_i)\}_{i=1}^N$ together with the observation $(x_{N+1},\hat{y}_{N+1})$ and a query for the next point $(x_{N+2},0)$. We repeat this process until we have obtained the requisite number of extrapolated values. We then computed the mean square error of the model extrapolation with the true value of the function at the corresponding 10 points.

We can see that the omnibus effect of introducing either the finite window length or the relational transformer is a significant improvement in performance (compare second and third rows of table with first). However, the breakdown according to curve types is quite different.  In particular, the 1d transformer with Learned Window attains poor performance on linear curves, but compensates with large improvements for sines and RBF curves, compared to the baseline. This is somewhat at odds with psychological data suggesting that people can extrapolate lines more accurately than oscillations \cite{ciccione,kalish}. By contrast, the similarity transformer improves on both lines and sines compared to the baseline, while preserving the relative difficulty between them. Finally, the variant with both a learned window and relational bottleneck attains the best performance of all, suggesting that both biases together are helpful for accurately predicting simple functions, moreso than either on its own.

\begin{table*}
\centering
\caption{Extrapolation accuracy, mean square error. Values are mean and standard error, over 3 copies of each network. We show an average over all test curves, as well as broken down by curve type.\label{tab:mse_results_table}}

\begin{tabular}{lllll}
%\toprule
 &              all &             lin &             rbf &            sine \\
%\midrule
1d transformer                       &  $0.545\pm 0.060$ &  $0.109\pm 0.005$ &  $1.501\pm 0.164$ &  $0.213\pm 0.047$ \\
1d transformer, learned window         &  $0.415\pm 0.022$ &  $0.368\pm 0.045$ &  $0.810\pm 0.020$ &  $0.056\pm 0.005$ \\
relational transformer               &  $0.401\pm 0.017$ &  $0.073\pm 0.020$ &  $1.109\pm 0.044$ &  $0.163\pm 0.042$ \\
relational transformer, learned window &  $0.365\pm 0.016$ &  $0.069\pm 0.026$ &  $1.063\pm 0.030$ &  $0.085\pm 0.014$ \\
%\bottomrule
\end{tabular}

\end{table*}

\begin{table*}
\caption{Estimated uncertainty results for the Relational Transformer model. Note that the baseline transformer model does not have any way to natively estimate uncertainty, so corresponding values are not shown. Values are mean and standard error, over 3 copies of each network. Optimal values are defined as in main text. \label{tab:std_results_table}}
\centering

\begin{tabular}{lllll}
%\toprule
 &               all &              lin &              rbf &             sine \\
%\midrule
relational transformer               &   $0.211\pm 0.019$ &   $0.068\pm 0.015$ &   $0.399\pm 0.028$ &   $0.241\pm 0.023$ \\
relational transformer, learned window &   $0.281\pm 0.050$ &   $0.092\pm 0.024$ &   $0.603\pm 0.135$ &   $0.238\pm 0.022$ \\
optimal & & .1 & .802 & .1
%\bottomrule
\end{tabular}

\end{table*}

\section{Uncertainty Estimation Results}
\label{std_results}
As mentioned previously, the relational transformer model has the property of natively estimating uncertainty of its own estimates. To evaluate this capability, we generate extrapolations out to $t=10$ steps as before, and average the predicted standard deviation at each step. We show in table \ref{tab:std_results_table} the results for the Relational Transformer, both with and without the learned window.

For the sake of having a comparison, we now consider the question of an optimal value for these estimates. For the cases of lines and sinusoids, we recall that they are generated using an underlying deterministic function corrupted by iid exogenous noise. A perfectly predictive and calibrated model, therefore, would be able to infer the underlying function, and would thus have an uncertain value that is equal to the exogenous noise, which in our case was $\sigma=.1$. The RBF functions are slightly different, because even in the absence of exogenous noise, the functions are sampled from a distribution rather than generated according to a deterministic formula, and thus they have an irreducible amount of unpredictability. However we can still define an optimal uncertainty estimate using the underlying kernel of the RBF process. More precisely, we define the optimal uncertainty estimate $\sigma_t$ to be the standard deviation of the RBF posterior distribution of $y_{t+N}|y_1,\dots, y_N$. For direct comparison with the models, we average these standard deviation values over $t=1,\dots, 10$ to obtain the value in the table. It is a mathematical fact that the posterior variance does not in fact depend on observed values $y_1,\dots, y_n$ of the function \cite{lucas}, and consequently all RBF curves have the same optimal uncertainty value.
 
In this case, the uncertainty estimates for the similarity transformer with and without the masking are essentially indistinguishable, for the cases of lines and sines. Both tend to underestimate the uncertainty for lines, and overestimate for sinusoids. We thus see the same order-of-difficulty effect as in the MSE values. Interestingly, the similarity transformer significantly underestimates the variance of RBF curves compared to the version with the masking. Thus we see another benefit to having both biases in the model, rater than just the Similarity Bottleneck. 

\section{Related work}
\label{related_work}
\subsection{Scalar Function Learning}
\label{fn_lrn_related_work}
Scalar function learning has been a classic topic in cognitive psychology \cite{delosh,mcdaniel,bott}, and has recently gained popularity as a test case for large neural network models as well. 

Modern modeling approaches of function learning are typically based on Gaussian Processes \cite{schulz,wilson,lucas} and autoregressive linear models \cite{maxent}. Suggestively, the Gaussian Process approach relies very heavily and explicitly on the notion of a similarity matrix, namely the covariance kernel of the process. There, however, the usage of the matrix conceptually differs somewhat from its usage in the present work, in that the matrix is there assumed to have a known parametric form, rather than constructed from the input data as we do. 

The general simplicity/tractability of the space and grounding in psychological data have also made it appealing as a test bed for neural networks. For example, in \cite{tmlr}, the authors used scalar function learning tasks analogous to those from the psychological literature to analyze a variety of self-supervised learning models, as well as to build models that more closely match patterns of human behavior. Works such as \cite{garg,Oswald} have begun to systematically evaluate the function learning capabilities of transformers, however they have focused more on iid interpolation, and less on the extrapolation/prediction setting that concerns us.

%The similarity matrix construction also has some interesting parallels with other work. To expand upon this, recall the assumption ath a similarity function $S(\cdot,\cdot)$ as given. We make the simple observation that if $y$ is some value, and $t\mapsto x_t$ is a scalar function, then $t\mapsto S(y,x_t)$ is also a scalar function, which can be thought of as an \textit{augmentation} of x. Thus effectively the similarity matrix-based models construct a family of data augmentations, parametrized by the input function itself. Another way to look at the models is thus that they are trained to simultaneously predict the values of an ensemble of augmented functions. This is conceptually very closely related to \cite{tmlr} paper, which also used a family of data augmentations to construct a self-supervised objective for scalar function learning. The difference here is that our augmentations are determined on-the-fly by the input functino itself, while in that work the augmentations were sampled from a pre-specified distribution. 

%The machine learning topic of Neural Processes is also highly relevant, as it can be regarded as replacing a fixed Gaussian kernel with a learable function parametrized by a neural network\cite{cnp,cnp2,cnp3}. 

\subsection{Relational Reasoning with Neural Networks}
\label{simmat_related_work}
As outlined previously, our approach on the Relational Bottleneck is directly inspired by \cite{esbn,corelnet}, in which a form of abstract reasoning is attained by forcing a network to attend to strictly relational information, encoded as a pattern of similarities between a sequence of inputs to the network. 

The same tasks in the ESBN, as well as some generalization to more complex images have been addressed in the recent GAMR model \cite{vaishnav};however despite the similar objective, the mechanism of the model is quite different from ESBN and CoRelNet, relying instead on a learned visual attention policy.

The property of working with similarity and relational information has also been a key ingredient in many previous models, albeit not with the absolute separation imposed by CoRelNet and ESBN. 
For example, the transformer architecture itself \cite{vaswani} is built on a simple form of all-to-all attention determined by patterns of similarity among inputs. Models such as the Differentiable Neural Dictionary \cite{dnd} and Neural Turing Machine \cite{graves} also make use of similarity computations in a key way, namely as attention patterns to lookups in external memory. Another strategy is to impose strong priors on the pattern of attention weights by imposing a graph structure on the data and allowing each node to attend directly only to nodes that are connected by short paths along the graph \cite{gnn,gat,gdl} This technique often requires strong assumptions about graph structure, and may not generalize to continuous cases such as scalar functions. 

The general motivation behind the similarity matrix itself, namely of separating ``sensory" from "abstract" processing,  have also been employed in the context of neuroscience. For example \cite{tem}  achieves a form of flexible generalization in navigation tasks by enforcing a separate ``abstract" network whose processing is divorced from the explicit perceptual information coming into the network. This model has been argued to be formally equivalent to a basic transformer with a specific form of positional encoding and key lookup \cite{temt}.

\section{Limitations and Future Work}
\label{limitations}
%In this work we have intentionally chosen to analyze data that are simple and low-dimensional in character. While  we believe that the data used in these experiments are a valuable and worthwhile subject of study in their own right (as further elaborated upon in the discussion section), it is also true that they are simpler than the stimuli for most visual reasoning tasks, and it remains to be seen to what extent the benefits of the modifications we have made to the transformer architecture will show similar benefits in other abstract visual reasoning tasks. 

While we have tried to motivate our specific choice of similarity function from mathematical considerations, it is not clear whether this is the best possible choice. In future work, it will be interesting to explore whether a model would be able to learn an appropriate similarity function directly from data, rather than having it pre-specified. 

%Finally, while the similarity matrix construction gives a highly generaly and formally elegant method to define uncertainty estimates, we have seen that the quantiative results leave some room for improvement in future work.

Further, we recall that the current form of the model also requires an invertible similarity function in order to read out raw values. Thus signed distance is workable but unsigned is not. As not all similarity functions of interest have this property, it is an interesting direction for future work to relax this assumption while keeping the same ensembling property of the model.

\section{Discussion}
\label{discussion}

Inspired by abilities of recent Cognitive Science models (namely ESBN/CoRelNet, and MaxEnt Function Learning), we have built models to predict scalar functions that impose a bottleneck through computation of a relational matrix, and imposition an adaptive attention window. We proposed how to implement both of these biases in the context of a standard transformer model, and found that both individually improved the extrapolation performance of a transformer model on scalar functions, with the greatest gain coming in a model that incorporated both. Furthermore,we showed that our Relational Transformer method can naturally be extended to give uncertainty estimates, differentiating it from transformer models of this task such as \cite{garg}. 

Thus, we have shown that ideas from cognitive science can be profitably adapted to large deep learning models to improve performance on the kinds of tasks that people are good at. On the other hand, our work also generalizes and extends aspects of the cognitive models which it adapts. Firstly, while the ESBN and CoRelNet models were trained in a supervised multiple-choice setting, we have here extended the Relational Bottleneck property of these models to the case of a generative, self-supervised prediction objective. Secondly, in \cite{maxent}, the window length parameter $L$ is treated as a descriptive hyperparameter, and is not given a principled normative account. In our Learnable Window implementation, by contrast, we effectively promote $L$ to a learnable parameter which can be optimized alongside the rest of the model in an end-to-end fashion. In future work, it would be extremely interesting to consider the $L$ of the learned model with those estimated from data on people.  

\section{Acknowledgements}
This project was supported by a Vannevar Bush Faculty Fellowship from the Office of Naval Research (ONR N00014-22-1-2002). SS is also supported by a T32 Training Grant in Computational Neuroscience (T32MH065214).  

\bibliographystyle{apacite}

\setlength{\bibleftmargin}{.125in}
\setlength{\bibindent}{-\bibleftmargin}

\bibliography{CogSci_Template}

\end{document}